\documentclass[conference]{IEEEtran}
 
\usepackage{graphicx}
\usepackage{amsmath}
\usepackage{amssymb}
\usepackage{booktabs}
\usepackage{hyperref}
\usepackage{geometry}
\usepackage{float}
\usepackage{array}
\usepackage{placeins}
\usepackage{seqsplit}
\usepackage{times}
\usepackage{setspace}
\usepackage{microtype}
\usepackage{tabularx}   
\usepackage{adjustbox}  
\usepackage{stfloats}   
\usepackage{lettrine}
\hypersetup{
    colorlinks=true,
    linkcolor=blue,
    citecolor=blue,
    urlcolor=blue,
}
 
\title{Does Federated Training Degrade Explainability?
A Comparative Analysis of Grad-CAM Faithfulness in Centralized vs. Federated Deepfake Detection}
 
\author{
    Muhammad Uzair Shahid \\
    \and
    Muhammad Zabil Mehboob \\
}
 
\date{}
 
\begin{document}
 
\twocolumn[
\begin{@twocolumnfalse}
    \maketitle
\end{@twocolumnfalse}
]

\begingroup
\renewcommand{\thefootnote}{}
\footnotetext[0]{%
    \textit{(Muhammad Uzair Shahid and
    Muhammad Zabil Mehboob are co-first authors.)}
    \textit{(Corresponding author:
    Muhammad Zabil Mehboob.)}

    \par
    Muhammad Uzair Shahid is an independent researcher
    based in Pakistan
    (e-mail: uzairshahid6760@gmail.com).

    \par
    Muhammad Zabil Mehboob is an independent researcher
    based in Pakistan
    (e-mail: m.zabil.mehboob@gmail.com).
}
\endgroup

\noindent\hspace{1em}\textbf{\textit{Abstract—}}\textbf{Automated detection has become the essential safeguard against realistic synthetic media, but deploying it at scale requires training on sensitive data from multiple organizations, putting privacy and the transparency demanded by forensic review in direct conflict. Federated learning resolves the privacy side by sharing only model updates rather than raw data, but whether this comes at the cost of trustworthy gradient-based explanations, which analysts rely on to decide whether or not to trust a flagged clip, has not been studied. We present the systematic study of Grad-CAM explanation faithfulness under federated deepfake detection, evaluating an EfficientNet-B4 detector on FaceForensics++ across five federated clients under varying degrees of data heterogeneity. We find that the relationship between federation and explanation quality is non-monotonic: at near-IID (Independent and Identically Distributed Data) and moderate heterogeneity, federated models match or exceed the faithfulness of a centralized baseline, but at extreme heterogeneity, explanations break down entirely, becoming spatially incoherent and, on average, causally misleading. This degradation is not a general cost of federated training but a threshold effect tied to client heterogeneity, and we offer practical guidance for identifying when privacy-preserving deepfake detection can be deployed safely and when its explanations can no longer be trusted. Code and data partitions are available for reproducibility.}

\vspace{0.5em}

\noindent\hspace{1em}\textbf{\textit{Impact Statement—}}\textbf{Federated learning offers a practical way to train deepfake detectors across organizations without requiring them to share sensitive media. However, privacy-preserving training alone does not establish that the resulting explanations remain trustworthy. This work advances understanding of this issue by systematically examining the faithfulness of Grad-CAM explanations under different levels of client heterogeneity. Our results show that federated training does not inherently reduce explanation faithfulness: under near-IID and moderate heterogeneity, federated models can maintain or exceed the explanation faithfulness of centralized training. In contrast, extreme heterogeneity can cause explanations to become spatially incoherent and causally misleading. These findings identify client heterogeneity as an important factor in evaluating the reliability of explainable federated AI systems, rather than treating federated training itself as the source of explanation degradation. The study therefore provides a practical basis for assessing when federated deepfake detection can support applications where both data privacy and trustworthy model explanations are required.}

\vspace{0.5em}

\noindent\hspace{1em}\textbf{\textit{Index Terms—}} \textbf{Area Over the Perturbation Curve (AOPC), Deepfake Detection, Explainable Artificial Intelligence (XAI), Explanation Faithfulness, FaceForensics++, Federated Learning, Grad-CAM (Gradient-weighted Class Activation Mapping), Non-IID Data (Non-Independent and Identically Distributed Data).}

\vspace{1em}
 
\section{Introduction}
 
\subsection{Motivation}
 
\lettrine[lines=2]{S}{ynthetic} media, also called deepfakes, has reached a level of realism that makes manual inspection unreliable and threatens digital forensics \cite{rossler2019faceforensics}, information integrity, and public trust. Convolutional neural networks trained on large benchmarks such as FaceForensics++ now classify manipulated facial video with high accuracy \cite{rossler2019faceforensics}, and therefore automated detection has become the front line against this threat. 
 
However, putting that detection into real forensic pipelines runs into two demands that pull in different directions. The first is privacy: training on sensitive user-generated content pooled from social media platforms, news agencies, and government bodies raises data-governance concerns that rule out naive centralized aggregation. The second is explainability: reviewers in high-stakes settings, such as content moderation, legal proceedings and journalism verification, cannot act on a bare binary output. They need explanations that point to the specific facial regions that drove a manipulation decision \cite{selvaraju2017gradcam, samek2017evaluating}.
 
Federated learning (FL) solves the first problem by letting multiple data holders train one model together without any of them sharing raw samples \cite{mcmahan2017communication}. Under Federated Averaging, each participant trains locally on their own private data and sends only parameter updates to a central server, which combines them into a global model through sample-weighted aggregation. Digital forensics, healthcare, and finance have picked up this approach for exactly that reason. 
 
This privacy-preserving setup can, however, affect what the model learns. Clients typically hold different mixes of manipulation types, and so moving from centralized to federated training can cause gradient averaging to pull client weights apart, changing the model’s internally learned features \cite{zhao2018federated}. Grad-CAM and similar post-hoc explanation methods \cite{selvaraju2017gradcam} read visual saliency straight off those gradients, so if federated aggregation distorts the gradient space, the resulting heatmaps may no longer track what the model is actually doing. Forensic analysts routinely trust a flagged clip because its heatmap looks plausible, yet this is precisely the signal that federated aggregation could compromise. 
 
\subsection{Research Gap}
 
While federated learning, deepfake detection, and explainable AI have each cultivated extensive bodies of literature, these domains have largely evolved in isolation.
 
Federated learning research focuses on convergence guarantees, communication cost, and classification accuracy under statistical heterogeneity \cite{mcmahan2017communication, zhao2018federated, li2020fedprox, li2020convergence, li2022niid}. The quality of the gradient-based explanations that come out of the aggregated model is rarely mentioned.
 
Deepfake detection research reaches state-of-the-art accuracy with centralized training pipelines and checks visual explanations such as Grad-CAM and SHAP only under centralized conditions \cite{rossler2019faceforensics, wang2025deepfake, mansoor2025explainable, sunkari2024deepexplain, maheshwari2024xaiart}. Federated deployment of these same detectors is not part of that conversation.
 
While research on XAI faithfulness has yielded robust quantitative tools—such as AOPC \cite{samek2017evaluating}, parameter-randomization sanity checks \cite{adebayo2018sanity}, and saliency reliability analyses \cite{kindermans2019unreliability}—these metrics are almost exclusively applied to centralized models. To date, they have not been adapted for federated settings, much less for visual forensic tasks.
 
Federated XAI does exist in a few other domains: supply chain credit assessment \cite{shi2025transxfed} and network intrusion detection \cite{bilal2025federated}. These studies show that explanations can be produced without exposing client data, but they work in non-visual settings and never quantitatively compare centralized against federated explanation faithfulness in visual deepfake detection.

\subsection{Hypothesis}
 
Our primary hypothesis is:
 
\textbf{H1:} Federated training introduces gradient inconsistency across client models that reduces the spatial faithfulness of post-hoc gradient-based explanations (Grad-CAM) relative to centralized baselines, and this degradation correlates monotonically with the degree of data heterogeneity across clients, controlled here through the Dirichlet concentration parameter $\alpha$.
 
Under H1, the most heterogeneous federated settings (low $\alpha$) should produce the least faithful explanations, while near-IID conditions (high $\alpha$) should come closest to centralized explanation quality.
 
\subsection{Contributions}
 
This paper contributes the following:
 
\begin{enumerate}
\item \textbf{Empirical measurement of XAI faithfulness degradation in federated deepfake detection.} We compare AOPC and Explanation IoU between a centralized EfficientNet-B4 baseline and federated counterparts across four controlled heterogeneity levels on FaceForensics++.
 
\item \textbf{A characterization of how heterogeneity and faithfulness relate.} We show that federation's effect on explanation quality is non-monotonic and confined to the extreme heterogeneity regime ($\alpha = 0.1$), backed by a perfect Spearman rank correlation ($\rho = -1.0$) between the Dirichlet concentration parameter and AOPC degradation.
 
\item \textbf{An open-source experimental framework.} All data partitions, training configurations, Grad-CAM maps, and evaluation scripts are released so the results can be replicated directly and reused as a foundation for federated XAI research in visual forensics.
 
\item \textbf{Practical guidance for practitioners.} We give concrete thresholds for when federated training preserves explanation quality and when it needs help, for instance from proximal regularization or explanation-aware aggregation.
\end{enumerate} 

\section{Related Work}

The proposed study sits at the intersection of four research areas that have largely developed on separate tracks: federated learning under statistical heterogeneity, deepfake detection, explainable artificial intelligence (XAI), and quantitative evaluation of explanation faithfulness. Federated learning research is centered on optimization efficiency and convergence. Deepfake detection work is centered on classification accuracy. XAI research examines explanation generation and reliability under centralized training. Federated XAI applications remain confined to low-dimensional domains. This section reviews each pillar in turn, then synthesizes the gap that motivates the present study.

\subsection{Federated Learning and Non-IID Data}

Federated Learning (FL) allows collaborative model training without exchanging raw data. Participating clients perform local optimization and share only model parameters, which makes FL suitable for privacy-preserving machine learning in healthcare, finance, mobile computing, and digital forensics. McMahan et al.~\cite{mcmahan2017communication} established the foundation of modern federated optimization with Federated Averaging (FedAvg): each client runs several local stochastic gradient descent (SGD) updates before sending parameters to a central server for weighted aggregation. Their experiments showed that FedAvg cuts communication rounds by an order of magnitude relative to synchronized distributed SGD while matching its classification performance. Their evaluation, however, is limited to communication efficiency and predictive accuracy; the quality and reliability of the learned gradient representations is left unexamined.

Subsequent work identified statistical heterogeneity as one of the primary obstacles to federated optimization. Zhao et al.~\cite{zhao2018federated} showed that when clients hold highly skewed local datasets, the global model's performance degrades substantially. They traced this to weight divergence: discrepancies between local and global label distributions produce increasingly inconsistent parameter updates over training, with accuracy reductions approaching 55\% under severe non-IID settings and divergence closely tied to the Earth Mover's Distance (EMD) between client and global data distributions..

Li et al.~\cite{li2020fedprox} proposed FedProx to address heterogeneous environments, adding a proximal regularization term to each client's local objective that discourages large deviations from the current global model. Unlike FedAvg, FedProx addresses both statistical and system heterogeneity and produces more stable convergence across clients with different computational capacities and data distributions, improving performance by approximately 22\% in highly heterogeneous settings. Constraining local optimization this way changes the gradient behavior used during backpropagation, and since post-hoc explainability methods derive explanations directly from these gradients, the effect of proximal regularization on explanation quality is still uninvestigated.

Li et al.~\cite{li2020convergence} provided the first rigorous convergence guarantees for FedAvg under non-IID data, proving an $O(1/T)$ convergence rate for strongly convex objectives and showing that increasing local optimization epochs introduces a bias proportional to the degree of statistical heterogeneity unless a decaying learning rate is used. This is optimization theory, but it carries a direct implication for explainability: gradient-based explanation methods assume gradients faithfully represent the model's decision process, so if federated optimization systematically biases those gradients, explanation faithfulness should degrade as well.

Li et al.~\cite{li2022niid} introduced NIID-Bench, a benchmark evaluating multiple federated algorithms across several forms of statistical heterogeneity --- label skew, feature skew, and quantity imbalance. No single algorithm consistently outperformed the others across all scenarios, and Dirichlet-based label partitioning has emerged as one of the most challenging and widely adopted methods for simulating heterogeneous client distributions. The Dirichlet concentration parameter ($\alpha$) has since become the standard mechanism for controlling statistical heterogeneity in federated learning experiments, and the present study adopts this framework to analyze explanation degradation under varying degrees of client diversity.

Taken together, the federated learning literature shows that statistical heterogeneity alters optimization dynamics through conflicting client updates and gradient divergence, but existing evaluation stays largely within convergence speed and classification accuracy. How these altered gradient fields affect post-hoc explainability has received little attention, leaving a gap between federated optimization theory and explainable AI.

\subsection{Deepfake Detection and Explainability}

Advances in generative AI have made manipulated multimedia content more realistic and more accessible, which makes automated deepfake detection an increasingly important problem. Detection models now achieve strong classification performance, but forensic and security deployments need more than accurate predictions --- they need explanations that let human analysts verify model decisions.

Wang et al.~\cite{wang2025deepfake} surveyed modern deepfake detection techniques from a reliability perspective and identified several persistent challenges, including limited transferability across manipulation techniques, insufficient interpretability, and robustness issues. They emphasized the need for reliable detection models in real-world settings and identified interpretability and robustness as important directions for further research, but did not empirically investigate how federated training affects the faithfulness of explanations in visual deepfake detection.

Gradient-based and feature-based explanation techniques have been used to study transparency in deepfake detection directly. Mansoor and Iliev~\cite{mansoor2025explainable} applied network dissection to CNN-based detectors, including ResNet-50, VGG-16, and Inception V3, and analyzed the interpretable facial concepts learned by these models. The detectors achieved F1-scores between 0.80 and 0.90, while the network dissection analysis identified interpretable facial concepts across different CNN architectures. This demonstrates the potential of model-interpretation techniques for analyzing CNN-based deepfake detectors, but their framework evaluates models under centralized training and does not consider explanation behavior under privacy-sensitive distributed learning.

Sunkari and Srinagesh ~\cite{sunkari2024deepexplain} proposed DeepExplain, integrating Grad-CAM and SHAP into a CNN-LSTM architecture trained on the DeepFake Detection Challenge (DFDC) dataset. Their framework uses Grad-CAM to visualize regions that contribute to model predictions and SHAP to quantify the contribution of input features, providing insights into the model's decision-making process. The study argues that these explainability mechanisms can foster trust and improve understanding of DeepFake detection decisions. However, the framework is evaluated under centralized training, so whether explanation consistency survives distributed training across heterogeneous clients remains unclear.

Maheshwari and Paulchamy~\cite{maheshwari2024xaiart} introduced XAI-ART, combining adversarial robustness training with explainable AI to improve both detector resilience and transparency, reaching approximately 97.5\% classification accuracy and maintaining high recall under adversarial attacks while producing interpretable attention maps. This shows that robustness and explainability can coexist in centralized deepfake detectors, but it does not examine whether federated optimization affects the stability or faithfulness of those explanations.

Existing deepfake detection research demonstrates that Grad-CAM and related techniques improve transparency and user trust under centralized training, but each study assumes a coherent gradient space produced by centralized optimization. None examine whether gradient interference from federated aggregation changes the spatial consistency or faithfulness of explanation maps, which leaves a gap for privacy-preserving deepfake detection systems.

\subsection{XAI Faithfulness Measurement}

Recent XAI research has shifted from generating visual explanations to evaluating whether those explanations faithfully represent a model's decision process. Grad-CAM, proposed by Selvaraju et al.~\cite{selvaraju2017gradcam}, remains the most widely used explanation technique for CNNs, producing heatmaps from class gradients, but it assumes a coherent gradient space that may not survive federated aggregation.

Samek et al.~\cite{samek2017evaluating} introduced the Area Over the Perturbation Curve (AOPC) to assess explanation quality objectively, measuring the drop in model confidence as important regions are progressively removed. Adebayo et al.~\cite{adebayo2018sanity} and Kindermans et al.~\cite{kindermans2019unreliability} showed that saliency maps need validation through sanity checks and robustness analysis, since visually plausible explanations do not always reflect actual model behavior.

These methods give explainability a reliable evaluation framework, but they have been validated almost exclusively under centralized training. Whether they apply to federated deepfake detection, where statistical heterogeneity may disrupt gradient consistency, remains largely unexplored.

\subsection{XAI within Federated Systems}

Federated XAI extends federated learning by integrating explainability into privacy-preserving machine learning. Existing work applies XAI to supply chain credit assessment \cite{shi2025transxfed} and intrusion detection \cite{bilal2025federated}, showing that explanations can be generated without exposing client data. 

These studies focus on non-visual applications, though, and do not examine whether explanation quality and faithfulness remain stable under federated training for visual deepfake detection. Quantitative comparisons between centralized and federated explanations for visual deepfake detection specifically have not been reported.

\subsection{Research Gap}

Table~\ref{tab:gap} summarizes where each research pillar stands and what it leaves open.

\begin{table*}[!t]
\centering
\caption{Summary of the Converging Literatures}
\label{tab:gap}
\begin{tabularx}{\textwidth}{@{} p{2.5cm} X X @{}}
\toprule
\textbf{Research Pillar} & \textbf{Established} & \textbf{Missing} \\
\midrule
Federated Learning~\cite{mcmahan2017communication, zhao2018federated, li2020fedprox, li2020convergence, li2022niid} &
FedAvg, FedProx, convergence analysis, weight divergence under non-IID data, Dirichlet $\alpha$ as the standard heterogeneity benchmark &
Effect of gradient divergence on explanation faithfulness remains unexplored \\
\addlinespace
Deepfake Detection~\cite{wang2025deepfake, mansoor2025explainable, sunkari2024deepexplain, maheshwari2024xaiart} &
High-performance detectors with Grad-CAM, SHAP, adversarial robustness, and centralized explainability &
Federated implementation and evaluation of explanation reliability \\
\addlinespace
XAI Faithfulness~\cite{selvaraju2017gradcam, samek2017evaluating, adebayo2018sanity, kindermans2019unreliability} &
Grad-CAM, AOPC, sanity checks, and saliency reliability provide objective explanation evaluation &
No application of these metrics to federated deepfake detection under statistical heterogeneity \\
\addlinespace
Federated XAI~\cite{shi2025transxfed, bilal2025federated} &
Explainability integrated into FL for credit assessment and intrusion detection &
High-dimensional visual forensics, quantitative comparison between centralized and federated explanations, and heterogeneity-aware faithfulness evaluation \\
\bottomrule
\end{tabularx}
\end{table*}

Federated learning, deepfake detection, XAI, and federated XAI have each advanced independently, but their intersection remains largely unexamined. Federated learning research has characterized gradient divergence and convergence behavior under non-IID data, but not its downstream effect on explanation faithfulness. Deepfake detection research has validated Grad-CAM and SHAP as explanation tools, but only under centralized training. XAI research has produced faithfulness metrics such as AOPC and sanity checks, but has not applied them to federated visual models. Federated XAI research has integrated explainability into FL for tabular and sequential domains such as credit assessment, intrusion detection, and speech, but has not extended to high-dimensional visual forensics or heterogeneity-aware faithfulness evaluation. This paper addresses that combined gap directly.

\section{Methodology}
 
\subsection{Dataset and Pre-processing}
 
All experiments used the FaceForensics++ (FF++) benchmark \cite{rossler2019faceforensics}, at the lightly compressed c23 encoding. The corpus contains 1{,}000 real videos sourced from YouTube and 1{,}000 videos for each of five manipulation methods: Deepfakes (DF), Face2Face (F2F), FaceShifter (FSh), FaceSwap (FS), and NeuralTextures (NT), for 6{,}000 clips across six classes. We collapse these to a binary real-vs.-fake label for classification.
 
\textbf{Frame extraction.} We sampled frames at a fixed temporal stride of 10 and localized face regions with the OpenCV DNN res10 SSD face detector (pretrained Caffe model, minimum detection confidence 0.50). Each detected face crop was resized to $224 \times 224$ pixels and saved as a JPEG at quality 95. The extraction pipeline is implemented in \textit{src\slash data.py}. An acceptance gate required at least 14{,}000 face crops before training could start.
 
\textbf{Dataset splits.} We partitioned the extracted crops into training (70\%), validation (15\%), and test (15\%) subsets using a fixed random seed (seed = 42) via NumPy's default random number generator, so the split is reproducible. The test split was fixed across all experiments and was never touched during architecture selection or hyperparameter tuning. It contains 12{,}037 images, confirmed against \textit{results\slash centralized\_metrics.json}.
 
\textbf{Normalization and augmentation.} Images were normalized with ImageNet channel statistics:
\begin{equation}
\text{mean} = [0.485,\ 0.456,\ 0.406]
\end{equation}
\begin{equation}
\text{std} = [0.229,\ 0.224,\ 0.225]
\end{equation}
Training images also received random horizontal flipping and mild color jitter (brightness, contrast, and saturation perturbations of $\pm 0.1$). Evaluation images received only resizing and normalization, with no stochastic augmentation.
 
\subsection{Federated Data Partitioning}
 
To simulate client heterogeneity, we partitioned the training corpus across $N = 5$ simulated clients with a Dirichlet distribution at concentration $\alpha$, following the non-IID partitioning protocol of Zhao et al.\ \cite{zhao2018federated}. For each class label, we drew a set of proportions from $\text{Dir}(\alpha \cdot \mathbf{1}_N)$ and assigned training indices to clients accordingly. Low $\alpha$ concentrates a client's data in one or two manipulation categories, approximating federated nodes that specialize in different forgery techniques. As $\alpha \to \infty$, the partition converges to IID.
 
We swept four heterogeneity levels: $\alpha = 0.1$, $0.3$, $0.5$, and $1.0$. At $\alpha = 0.1$, a client may draw the overwhelming majority of its samples from a single manipulation category. At $\alpha = 1.0$, client distributions closely approximate uniform coverage across all classes. All partitions were generated deterministically (seed = 42) and serialized to \textit{data\slash partitions\slash alpha\_*.json}.
 
\subsection{Model Architecture}
 
The backbone is EfficientNet-B4 \cite{tan2019efficientnet}, pretrained on ImageNet and fine-tuned for binary deepfake classification. We replaced the original classification head with:
\[
\text{GlobalAvgPool} \rightarrow \text{Dropout}(p{=}0.3) \rightarrow \text{Linear}(C \rightarrow 2)
\]
The EfficientNet-B4 feature extractor (\texttt{\seqsplit{backbone.extract\_features}}) produces spatial feature maps of shape $(B, C, 7, 7)$ at the penultimate convolutional stage; we globally average-pool these to a $(B, C)$ vector before the classification head. All backbone layers stayed unfrozen during fine-tuning. The model is implemented in \textit{src\slash model.py} and exposes a \texttt{\seqsplit{get\_last\_conv()}} accessor returning \texttt{\seqsplit{backbone.\_conv\_head}}, the Grad-CAM target layer.
 
We chose EfficientNet-B4 for its track record on FF++ and because gradient-based attribution methods apply directly to convolutional architectures. This work introduces no new architecture; its contribution is an analysis of how explanations behave under federation.
 
\subsection{Centralized Baseline Training}
 
The centralized baseline was trained on the full training split without client partitioning, using Adam (learning rate $10^{-4}$, batch size 32) with cross-entropy loss, for 10 epochs, with checkpointing at peak validation accuracy. The best checkpoint (\textit{models\slash centralized\slash best.pt}) was used for all subsequent Grad-CAM generation and AOPC evaluation, and serves as the performance and faithfulness upper bound against which we compare all federated conditions.
 
\subsection{Federated Training Protocol (FedAvg)}
 
We implemented federated training as a sequential FedAvg simulation in pure PyTorch (\textit{src\slash federated\_server.py}), with no external orchestration framework (e.g., Ray or Flower). Only one client model resides in GPU memory at a time, so memory use stays bounded regardless of client count.
 
\begin{figure}[t]
\centering
\includegraphics[width=0.9\linewidth]{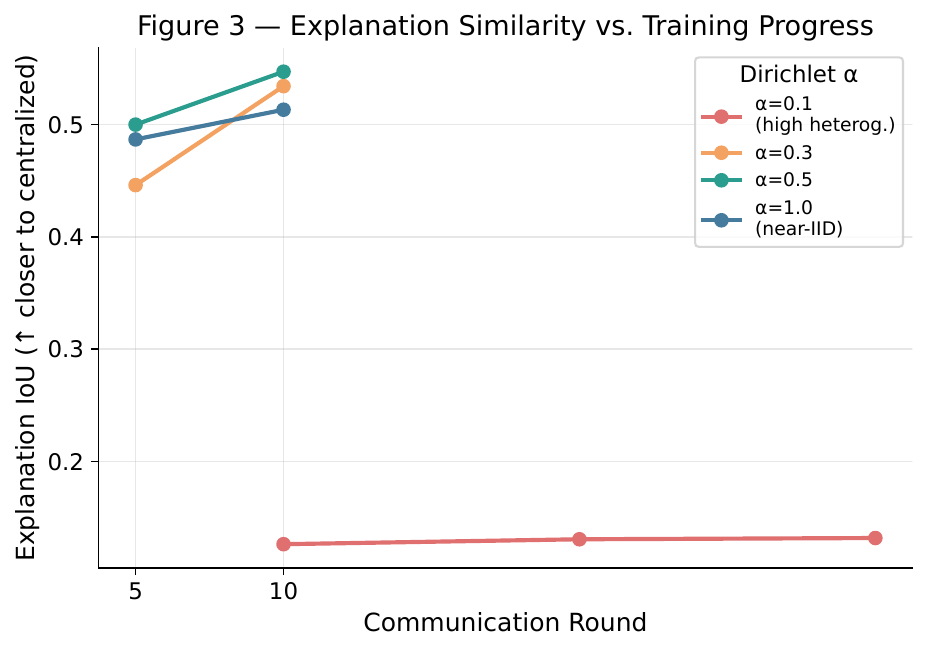}
\caption{Explanation IoU across communication rounds for $\alpha \in \{0.1, 0.3, 0.5, 1.0\}$.}
\label{fig:iou_rounds}
\end{figure}

\textbf{Communication rounds.} Given single-GPU compute constraints, we ran $\alpha \in \{0.3, 0.5, 1.0\}$ for 10 communication rounds, by which point both validation accuracy and Explanation IoU had plateaued (Fig.~\ref{fig:iou_rounds}). To separate slow convergence from genuine non-convergence at the most heterogeneous setting, we extended $\alpha = 0.1$ to 30 rounds; metrics stayed essentially flat over that extension (Expl. IoU: 0.127 at round 10 $\rightarrow$ 0.132 at round 30), which points to non-convergence rather than a transient slowdown.
 
\textbf{Local training.} Each round consisted of one local epoch per client, following the FedAvg protocol of McMahan et al.\ \cite{mcmahan2017communication}, with Adam (learning rate $10^{-4}$), cross-entropy loss, batch size 32, and mixed-precision training where CUDA was available.
 
\textbf{Aggregation.} After all five clients finished local training in a round, we updated the global model by sample-weighted averaging:
\begin{equation}
w^{t+1} = \sum_{k=1}^{K} \frac{n_k}{n} \, w_k^{(t+1)}
\end{equation}
where $n_k$ is the number of samples client $k$ processed in that round and $n = \sum_k n_k$. We saved checkpoints every 5 rounds; for $\alpha = 0.1$ we kept rounds 10, 20, and 30, and for $\alpha \in \{0.3, 0.5, 1.0\}$ we kept rounds 5 and 10.
 
\subsection{Grad-CAM Explanation Generation}
 
We applied Gradient-weighted Class Activation Mapping (Grad-CAM) \cite{selvaraju2017gradcam} to the centralized model and to every federated final-round checkpoint, using the \textit{pytorch-grad-cam} library. The target layer was \texttt{\seqsplit{backbone.\_conv\_head}} and the target output was the fake class (index 1). For each test image, the Grad-CAM engine computed the gradient of the target class score with respect to the convolutional feature map and applied a ReLU-weighted linear combination. The resulting CAM was a float32 array of spatial dimensions $(H, W)$, normalized to $[0,1]$ and saved as a \texttt{.npy} file indexed by image ID. All CAMs were generated with \texttt{shuffle=False} so image IDs stayed aligned across model configurations.
 
\subsection{Faithfulness Metrics}
 
\subsubsection{Area Over the Perturbation Curve (AOPC)}
 
AOPC \cite{samek2017evaluating} tests whether the regions a Grad-CAM heatmap highlights are causally important to the model's decision. For each test image and its CAM:
 
\begin{enumerate}
\item Record the baseline confidence $f_0$: the model's softmax probability for the fake class on the unperturbed image.
\item For each step $k \in \{1, \ldots, 10\}$, mask the top $k \cdot 10\%$ most salient pixels (ranked by CAM value) by replacing them with the per-channel image mean.
\item Record the model confidence $f_k$ on the masked image.
\end{enumerate}
 
AOPC is computed as:
\begin{equation}
\text{AOPC} = \frac{1}{10} \sum_{k=1}^{10} \left( f_0 - f_k \right)
\end{equation}
 
A higher AOPC means masking the highlighted region produces a larger confidence drop, i.e., a more faithful explanation. A negative AOPC means the highlighted region is not causally relevant to the prediction, or is actively misleading. We computed AOPC for every image in the held-out test set, for each model configuration.
 
\subsubsection{Explanation IoU}
 
To quantify spatial agreement between explanations from different model configurations, we binarized each CAM at its 80th percentile (marking the top 20\% of activations as salient) and computed Intersection-over-Union between the two binary masks:
\begin{equation}
\text{Expl. IoU} = \frac{|B_{\text{central}} \cap B_{\text{fed}}|}{|B_{\text{central}} \cup B_{\text{fed}}|}
\end{equation}
This metric serves two purposes: a cross-model similarity score between centralized and federated explanations at convergence, and a within-run convergence monitor across communication rounds.
 
\subsection{Statistical Testing}
 
To check whether AOPC differences were statistically significant rather than incidental, we applied the following tests, all implemented in \textit{src\slash metrics.py} with \textit{scipy.stats}:
 
\begin{itemize}
\item \textbf{Mann–Whitney U test} (one-sided; $H_1$: central AOPC $>$ federated AOPC): a non-parametric rank test suited to non-normal AOPC distributions across the full test set, applied per $\alpha$ condition against the centralized baseline.
\item \textbf{Cohen's $d$} (pooled standard deviation): quantifies effect size alongside the U-test $p$-values.
\item \textbf{Spearman rank correlation ($\rho$)} between the four $\alpha$ values and the corresponding mean AOPC degradation ($\Delta$ = central mean AOPC $-$ federated mean AOPC), testing whether degradation orders monotonically with data heterogeneity.
\end{itemize}
 
\section{Results}
 
We organize the results around the sub-questions from the research proposal. Table~\ref{tab:results} reports the aggregated performance and faithfulness metrics for all experimental conditions. All federated results for $\alpha \in \{0.3, 0.5, 1.0\}$ reflect the round-10 checkpoint; the $\alpha = 0.1$ result reflects round 30, after the extended run that confirmed non-convergence.
 
\subsection{Detection Accuracy }
 
\begin{table*}[!t]
\centering
\caption{Detection accuracy, AUC, mean AOPC, and Explanation IoU across the centralized baseline and four federated $\alpha$ conditions.}
\label{tab:results}
\begin{adjustbox}{max width=\textwidth}
\begin{tabular}{lccccc}
\toprule
Setup & Accuracy (\%) & AUC & AOPC Mean & AOPC Std & Expl. IoU \\
\midrule
Centralized (baseline) & 97.31 & 0.9947 & 0.0694 & 0.3082 & --- \\
Fed. $\alpha = 1.0$ (near-IID) & 96.68 & --- & 0.1657 & 0.3482 & 0.513 \\
Fed. $\alpha = 0.5$ & 95.31 & --- & 0.1080 & 0.3005 & 0.547 \\
Fed. $\alpha = 0.3$ & 94.43 & --- & 0.0492 & 0.2827 & 0.534 \\
Fed. $\alpha = 0.1$ (30 rounds) & 78.42 & --- & $-0.0104$ & 0.0555 & 0.132 \\
\bottomrule
\end{tabular}
\end{adjustbox}
\end{table*}
 
The centralized baseline reached 97.31\% accuracy and an AUC of 0.9947 on the 12{,}037-image held-out test set, setting a strong performance ceiling. Accuracy fell monotonically as heterogeneity increased (Table~\ref{tab:results}). At $\alpha = 1.0$ (near-IID), accuracy dropped only to 96.68\%, within the 3--8\% tolerance the research proposal anticipated and consistent with prior federated learning results. At $\alpha = 0.3$ and $\alpha = 0.5$, accuracy stayed within that range (94.43\% and 95.31\%). At the most extreme heterogeneity level, $\alpha = 0.1$, accuracy collapsed to 78.42\%, a drop of nearly 19 percentage points from the centralized baseline. FedAvg aggregation fails to reconcile highly divergent client updates at this extreme partitioning regime, a result that shapes how we read the corresponding XAI findings.
 
\subsection{Explanation Faithfulness}
 
\begin{figure}[t]
\centering
\includegraphics[width=1\linewidth]{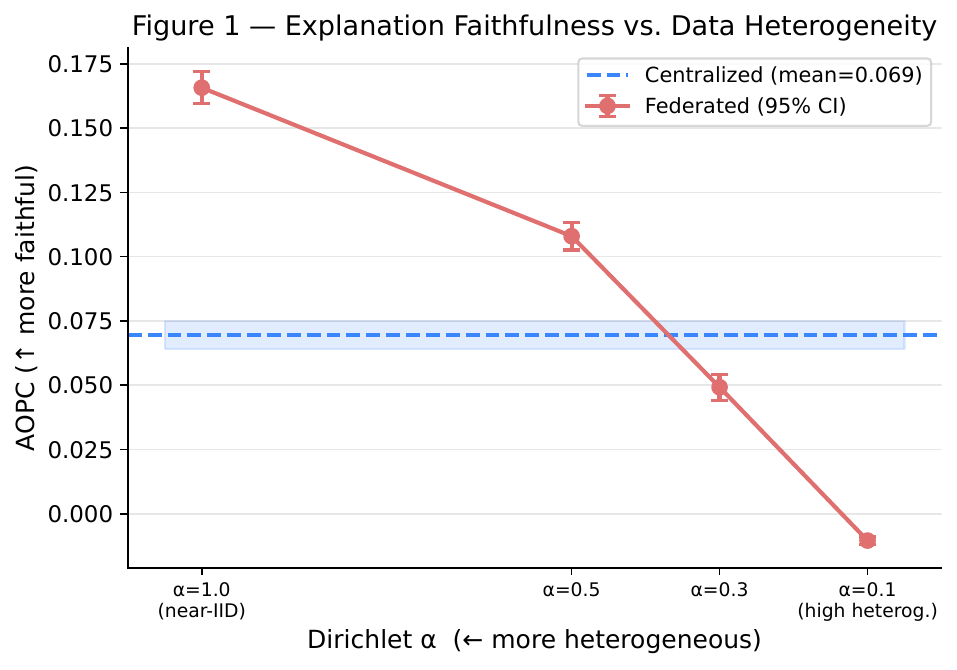}
\caption{Mean AOPC for each configuration (centralized, $\alpha = 1.0, 0.5, 0.3, 0.1$), with error bars showing the standard deviation.}
\label{fig:fig1_aopc_vs_alpha}
\end{figure}
 
The AOPC faithfulness metric follows a pattern that does not track heterogeneity in one direction. The centralized baseline's AOPC was 0.0694 ($\sigma = 0.3082$). Across the federated conditions:
 
\begin{itemize}
\item $\alpha = 1.0$: mean AOPC $= 0.1657$ ($\sigma = 0.3482$), a margin of $+0.0963$ over the centralized baseline.
\item $\alpha = 0.5$: mean AOPC $= 0.1080$ ($\sigma = 0.3005$), $+0.0386$ over baseline.
\item $\alpha = 0.3$: mean AOPC $= 0.0492$ ($\sigma = 0.2827$), $-0.0202$ below baseline.
\item $\alpha = 0.1$: mean AOPC $= -0.0104$ ($\sigma = 0.0555$), negative: the explanations are non-informative or actively misleading.
\end{itemize}
 
The one-sided Mann–Whitney U test ($H_1$: central AOPC $>$ federated AOPC) found the centralized model was not superior in faithfulness at $\alpha \in \{0.5, 1.0\}$ ($p = 1.0$ in both cases), with Cohen's $d$ of $-0.13$ and $-0.29$. These negative effect sizes mean that under near-IID and moderate heterogeneity, the federated models produced more faithful explanations than the centralized baseline --- a result that runs against the intuition that federation should only hurt explanation quality. The U-test reached significance ($p \approx 0$) only at $\alpha = 0.3$ ($d = 0.068$, small effect) and $\alpha = 0.1$ ($d = 0.315$, small-to-medium effect), where federated AOPC fell below the centralized reference.
 
These results partially support hypothesis H1. At low-to-moderate heterogeneity, federated training does not degrade explanation faithfulness and may modestly improve it. Only at extreme heterogeneity ($\alpha = 0.1$), where accuracy has already collapsed, does AOPC turn negative and faithfulness clearly fail.
 
\subsection{Heterogeneity Effect on Explanation Degradation }
 
The Spearman rank correlation between $\alpha$ and mean AOPC degradation ($\Delta$ = central mean AOPC $-$ federated mean AOPC) was $\rho = -1.0$ ($p < 0.001$): a perfect monotonic relationship between heterogeneity and the direction of AOPC change. Increasing $\alpha$ (lower heterogeneity) tracked with rising AOPC relative to baseline, while decreasing $\alpha$ tracked with deteriorating or negative AOPC. The perfect rank correlation ($|\rho| = 1.0$) confirms that AOPC degradation orders monotonically with heterogeneity, though the direction at $\alpha \in \{0.5, 1.0\}$ runs opposite to what the primary hypothesis predicted.
 
The relationship between federated heterogeneity and explanation quality is monotonic but not uniformly degrading. Both detection performance and faithfulness degrade primarily in the extreme heterogeneity regime ($\alpha = 0.1$); moderate-to-low heterogeneity preserves, or even improves, explanation faithfulness relative to the centralized model.
 
\subsection{Explanation Stability Across Communication Rounds}

\begin{figure}[t]
\centering
\includegraphics[width=0.9\linewidth]{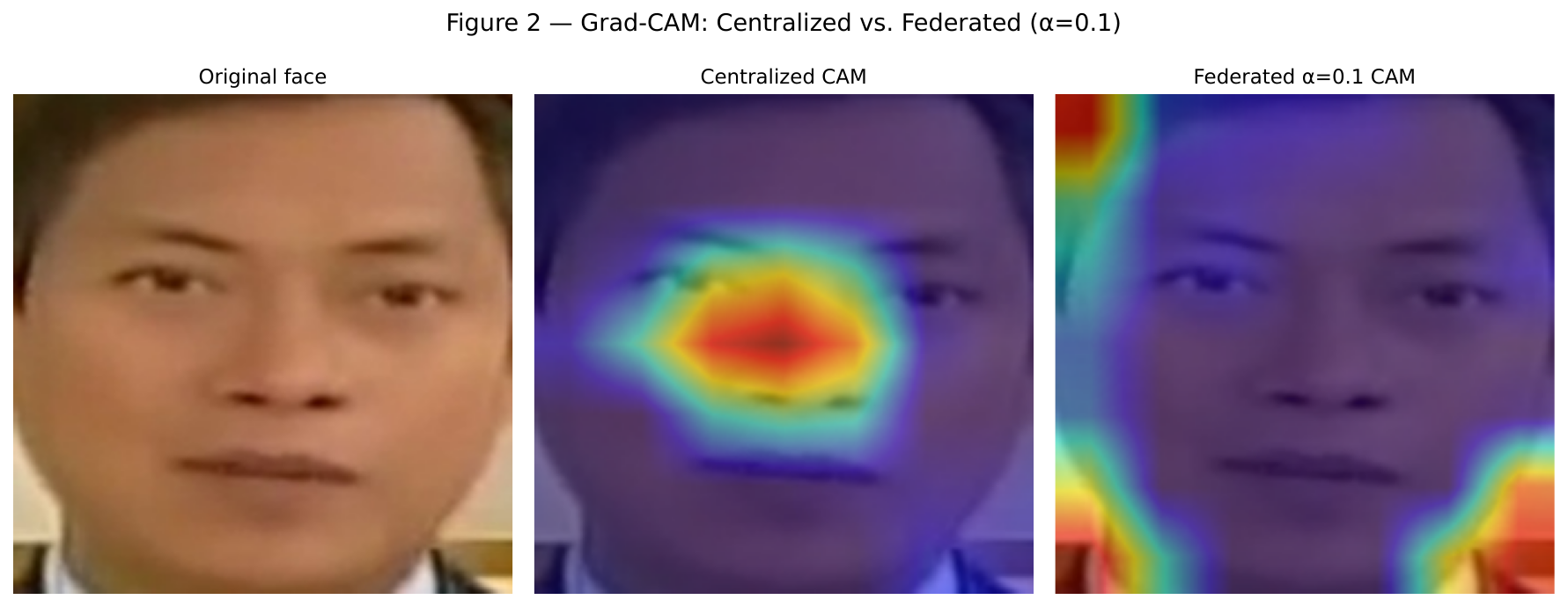}
\caption{Qualitative Grad-CAM comparison. The centralized baseline (center) yields coherent focal activations, whereas the federated model at extreme heterogeneity ($\alpha=0.1$, right) produces diffuse, spatially incoherent heatmaps.}
\label{fig:cam_compare}
\end{figure}
\FloatBarrier
 
We tracked Explanation IoU across communication rounds to check whether federated explanations stabilize over training:
 
\begin{itemize}
\item $\alpha = 1.0$: IoU $= 0.487$ at round 5 $\rightarrow 0.513$ at round 10.
\item $\alpha = 0.5$: IoU $= 0.500$ at round 5 $\rightarrow 0.547$ at round 10.
\item $\alpha = 0.3$: IoU $= 0.446$ at round 5 $\rightarrow 0.534$ at round 10.
\item $\alpha = 0.1$: IoU $= 0.127$ at round 10 $\rightarrow 0.131$ at round 20 $\rightarrow 0.132$ at round 30.
\end{itemize}
 
For $\alpha \in \{0.3, 0.5, 1.0\}$, Explanation IoU rose clearly between round 5 and round 10, showing that federated and centralized explanations aligned spatially as training advanced. In all three cases the trajectory had plateaued by round 10, which is why we stopped training there under single-GPU compute constraints.
 
For $\alpha = 0.1$, the trajectory looks different: IoU stayed low through the entire extended run, rising from 0.127 at round 10 to just 0.132 at round 30, a change of 0.005 over 20 additional rounds. This near-zero gradient does not match the gradual-convergence profile of a slowly learning model; it indicates that the global model failed to converge under extreme client divergence. The extended run rules out the alternative explanation that $\alpha = 0.1$ was simply slow to learn, and supports genuine non-convergence.
 
The qualitative heatmap comparison (Fig.~\ref{fig:cam_compare}) visually corroborates the quantitative findings. While our metrics indicate that models trained under moderate heterogeneity maintain spatial agreement with the baseline, the extreme heterogeneity condition ($\alpha = 0.1$) demonstrates a complete collapse in spatial coherence. As depicted in the figure, the centralized model focuses tightly on core facial features, whereas the federated model at $\alpha = 0.1$ disperses attention toward diffuse, peripheral regions. 
 
\subsection{Adebayo et al.\ Sanity Check}
 
Before faithfulness evaluation, the parameter-randomization test confirmed that the trained centralized model's Grad-CAM maps do not function as model-agnostic edge detectors: mean Explanation IoU between trained and fully weight-randomized CAMs fell below the 0.3 threshold set a priori, validating that the attribution maps are sensitive to the model's learned parameters.
 
\subsection{Summary of Findings}
 
\hspace{\parindent}\textbf{Accuracy.} Federated training degrades detection accuracy monotonically with data heterogeneity, from a 0.63 pp drop at $\alpha = 1.0$ to an 18.89 pp collapse at $\alpha = 0.1$. At moderate heterogeneity ($\alpha \in \{0.3, 0.5\}$), degradation stays within the 3--8 pp range prior federated literature anticipates.
 
\textbf{Explanation faithfulness.} The effect of federated training on AOPC depends on heterogeneity and is not uniform. At $\alpha \in \{0.5, 1.0\}$, federated models produce higher mean AOPC than the centralized baseline (not statistically worse; $p = 1.0$). At $\alpha = 0.3$, AOPC falls modestly below baseline (significant but small effect; $d = 0.068$). At $\alpha = 0.1$, AOPC turns negative ($d = 0.315$): the explanations stop being causally informative.
 
\textbf{Heterogeneity monotonicity.} Spearman $\rho = -1.0$ ($p < 0.001$) confirms a perfect rank-order relationship between data heterogeneity and AOPC degradation. This relationship implies that AOPC degradation is an extreme-heterogeneity phenomenon, not a general property of federated training.
 
\textbf{Convergence of explanations.} Explanation IoU rises across communication rounds for $\alpha \in \{0.3, 0.5, 1.0\}$, plateauing by round 10. For $\alpha = 0.1$, IoU stays flat across all 30 rounds, confirming non-convergence in this regime.

\section{Discussion}
 
\subsection{Federation Does Not Simply Break Explanations}
 
The primary result of this study was contrary to the expected outcome. Distributing training across heterogeneous clients was expected to weaken gradient-based explanations, but moderate heterogeneity instead improved them. Federated models at $\alpha = 1.0$ and $\alpha = 0.5$ produced mean AOPC values of 0.1657 and 0.1080, both above the centralized baseline of 0.0694. A Mann-Whitney U test found no advantage for the centralized model at either setting ($p = 1.0$), and negative Cohen’s $d$ values ($-0.29$ and $-0.13$) confirm the direction of the difference.
 
This pattern is unlikely to result from random sampling variation. The standard deviations remain within a comparable range (0.30--0.35), while the Explanation IoU values of 0.513 and 0.547 indicate that the federated heatmaps remain spatially consistent with the centralized ones. At the same time, they achieve higher scores on the causal AOPC criterion. These results suggest that, under near-IID and moderately heterogeneous conditions, federated training can improve explanation quality rather than simply preserve it.
 
One possible explanation is the implicit regularization introduced by federated aggregation. When five clients optimize on different subsets of the FaceForensics++ manipulation categories, each local gradient captures a slightly different view of the forgery decision boundary. Averaging these gradients gives the global model a form of gradient ensemble, which can reduce the client-specific activations that a single training run might otherwise overfit to. As a result, the model may focus more consistently on genuine forensic cues, such as boundary mismatches and texture irregularities, rather than dataset-specific correlations. Because Grad-CAM uses the predicted class gradients with respect to the final convolutional layer, more focused and causally relevant activations can result in higher AOPC scores. This mirrors the known effect of ensemble learning on explanation quality \cite{samek2017evaluating}, produced here implicitly through FedAvg aggregation rather than by design.
 
A second, compatible explanation points at the centralized baseline itself. Its AOPC of 0.0694 is modest even though the model generalizes well (97.31\% accuracy, AUC 0.9947). A model can reach near-perfect accuracy through a mix of solid and spurious features and still produce diffuse Grad-CAM maps that never concentrate on the regions that actually decide the outcome. Federated training, by narrowing each client's effective training distribution, may push the global model toward features that generalize across clients, and those are the features the gradient signal underlying Grad-CAM ends up attending to more tightly.
 
We do not claim to have settled why federated models score better on AOPC under moderate heterogeneity; that would require ablations beyond what this study set out to do. What the results do establish is that the effect is real and repeats across settings, and that it belongs in how practitioners weigh the trade-offs of federated deepfake detection.
 
\subsection{The Collapse at Extreme Heterogeneity Is a Different Kind of Failure}
 
The moderate-heterogeneity results are surprising in direction. The $\alpha = 0.1$ result is alarming in scale. Mean AOPC dropped to $-0.0104$: masking the regions Grad-CAM emphasizes increases the model's confidence in the fake class, on average. The highlighted regions are not just uninformative here. They actively point the wrong way. Explanation IoU collapsed to 0.132, and extending training to 30 rounds barely moved it (0.127 to 0.132 over 20 additional rounds). That near-flat trajectory rules out slow convergence as the explanation; this is non-convergence.
 
The mechanism is not subtle. At $\alpha = 0.1$, the data distribution across the five clients becomes highly skewed, with each client receiving samples mainly from one manipulation category. A client dominated by DeepFakes, for instance, will mostly learn patterns associated with that type of forgery, while a NeuralTextures-heavy client is exposed to a very different set of features. FedAvg then has to combine these conflicting local updates, making it difficult for the global model to learn a representation that works well across all manipulation types. This is reflected in the large drop in accuracy to 78.42\%, which is 18.89 percentage points below the centralized result. The explanation metrics point to a more serious issue: the negative AOPC and near-zero IoU indicate that the model is no longer producing a stable gradient structure. In other words, the problem is not only that the classifier becomes less accurate; its explanations also become unreliable. For a forensic analyst, a heatmap that does not consistently correspond to the model’s actual decision could be misleading, something that the accuracy numbers alone would not show.
 
The deployment lesson follows directly. Organizations using federated deepfake detection across highly specialized or domain-skewed clients, such as nodes focused on a single platform or compression type, should not rely on accuracy alone when deciding whether a model is ready for deployment. A model may still meet a minimum accuracy threshold under extreme heterogeneity even when its explanations have become unreliable. In such cases, presenting a Grad-CAM heatmap to a reviewer could give a false sense of confidence and potentially influence their judgment in the wrong direction.
 
\subsection{What the Monotonicity Result Tells Us}
 
The Spearman rank correlation between the Dirichlet concentration parameter $\alpha$ and mean AOPC degradation ($\Delta$ = centralized AOPC minus federated AOPC) is $\rho = -1.0$ ($p < 0.001$), indicating a perfectly consistent relationship across the four settings. As $\alpha$ decreases and client heterogeneity increases, the difference between centralized and federated AOPC becomes progressively larger in the unfavorable direction.
 
This finding points to more than a simple statistical relationship. It suggests that $\alpha$ can serve as an indicator of explanation reliability in this federated setting. If practitioners know the approximate Dirichlet concentration of their client data, or can estimate the Earth Mover’s Distance and its relationship with weight divergence \cite{zhao2018federated}, they can anticipate potential risks to Grad-CAM explanations. This also provides a practical way to assess data heterogeneity during onboarding and identify systems that may require additional evaluation before being used in workflows that depend on model explanations.
 
It is important to clarify what this monotonic relationship does not imply. It does not mean that every increase in heterogeneity causes the same degree of degradation; the results show that the effect is concentrated at the extreme end. The perfect rank ordering is driven in part by the sharp drop at $\alpha = 0.1$, while the remaining federated conditions ($\alpha \in {0.3, 0.5, 1.0}$) remain above the centralized baseline. A more accurate interpretation is that explanation faithfulness remains relatively stable across a broad range of heterogeneity levels before declining sharply beyond a tipping point.
 
\subsection{Practical Implications for Federated Forensic Pipelines}
 
Put together, the results give practitioners a workable set of guidelines.
 
At moderate heterogeneity ($\alpha \geq 0.3$), federated training does not compromise, and may strengthen, Grad-CAM faithfulness relative to a centralized baseline. Privacy-preserving federated deployment looks safe from an explainability standpoint here, and no special XAI mitigation is needed beyond standard sanity checks such as the parameter-randomization test \cite{adebayo2018sanity}.
 
At extreme heterogeneity ($\alpha \approx 0.1$), the model's explanations cannot be trusted. Options worth considering include proximal regularization such as FedProx \cite{li2020fedprox} to constrain client divergence; substantially more communication rounds, though our 30-round extension showed little payoff even then; explanation-aware aggregation that weights client contributions by how consistent their gradient directions are; or restricting human-in-the-loop review to the model's raw confidence score rather than a spatial heatmap that has lost its causal grounding.
 
Accuracy alone is not a sufficient gate. A model running at 78.42\% accuracy ($\alpha = 0.1$) would likely get caught by an accuracy threshold before deployment. But the statistically significant drop in AOPC observed at $\alpha = 0.3$, where accuracy stays at 94.43\%---comfortably within an acceptable operational range---shows that explanation degradation can begin before accuracy falls below deployment thresholds. AOPC, or a lightweight proxy for it, belongs in federated model evaluation pipelines as a routine check, not an afterthought.
 
\subsection{Limitations}
 
A few limits bound what we can claim here. The federated simulation runs sequentially on a single GPU with $N = 5$ clients, which does not capture the asynchrony, partial participation, or communication delays of a real federated deployment. We evaluate one backbone (EfficientNet-B4) and one post-hoc explanation method (Grad-CAM); whether the AOPC pattern generalizes to transformer-based detectors or attention-based explanations is still open. The Dirichlet partition is a controlled stand-in for heterogeneity; real federated nodes carry structured heterogeneity, platform-specific compression, device-dependent quality, that a Dirichlet model approximates without fully capturing. Our communication-round budget was limited by single-GPU resources, and the non-convergence seen at $\alpha = 0.1$ might in principle respond to more rounds or adaptive aggregation that we were not able to fully explore. Each of these points toward a natural follow-up.
 
\section{Conclusion}
 
This paper asked a question the literature had not answered directly: does federated training weaken the faithfulness of Grad-CAM explanations in deepfake detection, and by how much? The answer is neither a clean yes nor a reassuring no. It is conditional on data heterogeneity, and that conditionality has real consequences for how practitioners should think about deploying explanation-dependent forensic systems.
 
Under near-IID and moderately heterogeneous conditions ($\alpha \in \{0.3, 0.5, 1.0\}$), federated EfficientNet-B4 models trained on FaceForensics++ match or exceed the centralized model's explanation faithfulness by AOPC, while keeping detection accuracy within the range prior federated learning work would predict. Explanation IoU above 0.5 at these settings confirms that the federated heatmaps line up spatially with the centralized reference. Privacy-preserving federated deployment does not inherently damage the explanations human reviewers rely on, as long as client heterogeneity stays moderate.
 
The behavior changes sharply under extreme heterogeneity ($\alpha = 0.1$). Accuracy drops by nearly 19 percentage points, while Explanation IoU falls to 0.132. More importantly, mean AOPC becomes negative, indicating that the regions highlighted by the model are no longer informative and may instead be causally misleading. Extending training to 30 communication rounds did not resolve the issue; the minimal change in IoU between rounds 10 and 30 suggests that the problem is not simply slow learning but persistent instability. The Spearman rank correlation of $\rho = -1.0$ between $\alpha$ and AOPC degradation further shows a perfectly monotonic relationship, suggesting that data heterogeneity can serve as an early indicator of downstream explanation quality risk.
 
Our starting hypothesis, that gradient inconsistency from federated aggregation degrades Grad-CAM faithfulness monotonically with heterogeneity, is only partly right. The monotonic ordering holds. However, the direction of the effect at moderate heterogeneity contradicts our initial hypothesis: rather than merely mitigating expected degradation, federated models in this regime actually outperform the centralized baseline in explanation faithfulness. That revision matters in its own right, because it reframes the question from "does federation hurt explanations?" to "at what heterogeneity threshold does it start to hurt them, and what drives the benefit below that threshold?"
 
The broader methodological point is that accuracy alone is not sufficient for deciding whether a federated model is ready for deployment in explanation-dependent forensic applications. Explanation faithfulness can deteriorate in ways that standard performance metrics fail to capture, even when accuracy remains well above operational thresholds. For this reason, AOPC or a comparable causal perturbation metric should be included as a routine part of federated model validation.
 
A few directions follow naturally from here. Explanation-aware aggregation, where client contributions are weighted by how consistent their gradient directions are rather than by sample count alone, offers a principled way to protect explanation quality under heterogeneity. Adaptive communication strategies that extend training specifically when Explanation IoU has not plateaued could address the non-convergence problem at extreme heterogeneity without committing to a fixed round budget in advance. Extending this analysis to transformer-based detectors, where attention maps rather than convolutional gradients carry the explanatory signal, would test whether these patterns hold outside gradient-based attribution. We release our full experimental pipeline so these extensions are easy to build on.

\end{document}